\documentclass[conference]{IEEEtran}

\usepackage[dvips]{graphicx}
\usepackage{amsmath,amssymb}
\usepackage{algorithm}
\usepackage{algorithmic}
\usepackage{flushend}
\usepackage{multirow}
\usepackage[caption=false,font=footnotesize]{subfig}
\usepackage{epstopdf}
\usepackage[symbol*]{footmisc}
\usepackage{url}

\usepackage{color}

\begin{document}
\title{2D SLAM Quality Evaluation Methods}
\date{}

\author{
\IEEEauthorblockN{Filatov Anton, Filatov Artyom, Krinkin Kirill}
\IEEEauthorblockA{Saint-Petersburg Electrotechnical University ``LETI''\\ St. Petersburg, Russia \\ {ant.filatov, art32fil}@gmail.com, kirill.krinkin@fruct.org\\}
\and
\IEEEauthorblockN{Chen Baian, Molodan Diana}
\IEEEauthorblockA{Massachusetts Institute of Technology\\  Cambridge, USA \\ {baian, molodan}@mit.edu\\}}
\maketitle

\begin{abstract}
SLAM (Simultaneous Localization and mapping) is one of the most challenging problems for mobile platforms and there is a huge amount of modern SLAM algorithms. The choice of the algorithm that might be used in every particular problem requires prior knowledge about advantages and disadvantages of each algorithm. This paper presents the approach for comparison of SLAM algorithms that allows to find the most accurate one. The accent of research is made on 2D SLAM algorithms and the focus of analysis is 2D map that is built after algorithm performance. Three metrics for evaluation of maps are presented in this paper.
\end{abstract}

\section{Introduction}

SLAM is a part of navigation problem for mobile platforms that involves capturing the data from sensors and then simultaneously creating map and determining their location on this map. Tasks that use SLAM are widespread nowadays and there exists a huge amount of algorithms that solve problems of simultaneous localization and mapping. They vary by types of sensors, ways of storing data, math apparatus for handling scan data, etc. Comparing the available algorithms and choosing the one best suited for a given environment is a challenge unto itself.

To estimate a SLAM algorithm quality means to estimate the accuracy of it's results that may be determined by the map and trajectory of the mobile platform that were built during the run. The easiest way to compare results of several algorithms on the same data sequence is to calculate each difference between the built map and the ground truth map (i.e. the absolute one). However, ground truth maps are not available for many valuable datasets. For these data sequences, even when they have accompanying ground truth trajectories, it is complicated to extract a map against which the map built by the SLAM algorithm can be compared. If the data set has no ground truth, one must estimate the quality of an algorithm's results by other means. To the best of authors knowledge, there are few such methods available.

The goal of this paper is to present methods for quantitative evaluation of 2D laser SLAM algorithm quality. The object of consideration is a result map, and by its visually assessing, it is possible to find out which algorithm creates the best map, i.e., to determine which map contains the lowest amount of noise, the most accurate walls, the lowest amount of artifacts etc. Some metrics of quantitative estimation are presented in this paper. 

The framework that evaluates the maps is also presented in this paper. The maps that are built with each considered algorithm are compared using several metrics and a score is assigned for each run. In this way, it is possible to a ranking of SLAM algorithms even if the ground truths for are not provided for the considered datasets.

Properties of algorithms such as performance speed or map resolution were not considered because the main goal was to judge the accuracy of the output. Thus, all of the algorithms had enough resources and were set up with the same map resolution.

The paper is structured as follows: the state-of-art in SLAM algorithm evaluation is presented in section II, section III contains the description of metrics that are used for algorithms evaluation, section IV describes the framework for estimation, and in section V one can find experimental results for considered SLAM algorithms.

\section{State-of-Art}

The evaluation of SLAM algorithms has always been an important but challenging problem. The most trivial and accurate way to evaluate an estimated map produced by SLAM algorithm quantitatively is by using distance from the ground truth map as in \cite{santos_ieee_ssrr_2013}. This should be the best possible way to evaluate an estimated map in principle, except for the fact that in practice, the expectation of estimated maps may slightly vary. Using different distance criteria, the evaluation metrics may favor a sharper or a more conservative map. Other features like favoring either false positive or false negative results could also be reflected by the selection of a proper criterion. As an example, in \cite{santos_ieee_ssrr_2013}, the selected criterion is normalized distance in terms of k-nearest neighbor. 

However, this trivial but accurate and straightforward approach is usually much less feasible in real life cases, where ground truth maps are really hard to obtain. A commonly used alternative, as seen in \cite{huletski_fruct_2015}, is to compare estimated trajectory with ground truth trajectory. In \cite{huletski_fruct_2015}, root-mean-square error (RMSE) is selected as the criterion for this metric. Compared to a ground truth map, a ground truth trajectory is much easier to obtain in practice, so that even though it is not as straightforward as using a ground truth map to evaluate the estimated map, comparison between the estimated trajectory and the ground truth trajectory is actually one of the most widely used metrics in real life situations.

Some further investigations are also brought up in this direction. In \cite{kummerle_auto_robots_2009}, the shortcomings of using a global reference frame in evaluation (e.g. penalizing error in the beginning of trajectory more than at the end) is discussed, and an alternative solution that is unbiased in this domain and only uses relative information is proposed. In \cite{funke_bmvc_2009}, to make the evaluation of a ground truth map more straightforward, an evaluation framework for SLAM algorithms is built using a ground truth trajectory and scan data to generate a "pseudo ground truth map". These generated maps may not be perfect, but this technique also successfully overcomes the difficulty of obtaining ground truth maps in real life situations.

However, in practice, even the existence of a ground truth trajectory is not always guaranteed. In real life application (instead of evaluation), as mentioned in \cite{cadena_ieee_tr_robot_2016} and \cite{jaulmes_cont_arch_robot_2009}, the importance of some possible awareness of failure and artifacts is also emphasized, while to our best knowledge no quantitative solution to achieve this awareness has been proposed. Though in \cite{burgard_ieee_iros_2009}, a quantitative way of comparing SLAM algorithms without perfect ground truth is proposed, the evaluation framework still requires manual work by humans for every  map. 

It is worth mentioning that except for map accuracy, the quality of SLAM algorithms is also evaluated against other features, including trajectory accuracy measured by global or relative error \cite{huletski_fruct_2015,kummerle_auto_robots_2009}, efficiency measured by convergence speed within different parameter setup \cite{li_yac_2017}, robustness against noise or outliers measured by sensitivity \cite{li_yac_2017}, as well as robustness among different sensors and sequences \cite{huletski_fruct_2015}. Though our testing service supports more comprehensive analysis, the metrics proposed in this paper are focused on map evaluation.

\section{Metrics} \label{sec:Metrics}
If a human were to be presented with a map generated by a SLAM algorithm, it is highly possible that he could determine the quality of the image representation of the map correctly. He could pick out superposed copies of the same room, crooked walls, unexpected lines in the middle of a room, etc. The quantitative estimation of a map could rely on the same idea: to extract all these features of each map and calculate their amount. This section describes several metrics that can help to determine the quality of a map without relying on a ground truth map or trajectory. For analysis one shouldn't be content with one separate metric because it might show only approximate results under special conditions. Thus as more metrics are used for analysis than more accurate conclusion may be drawn.

In the descriptions that follow, the format of the maps that are being processed is a 2D matrix represented by an occupancy grid where each cell is assigned a probability of being occupied. The free cells in the image representation are defined as white, while the higher the probability that a cell is occupied, the darker it is in the image. By default, the color of an unknown cell in the image representation is also white, since this makes it easier to identify the failures and artifacts in this setup. 

The following metrics are presented to evaluate the quality of a SLAM algorithm:
\begin{itemize}
    \item The proportion of occupied and free cells, which allows one to determine whether walls on a map are blurred and to check if there are extra walls that appeared because of a failure of algorithm;
    \item the amount of corners in a map, which measures the precision of a map. A map of low quality should have more corners than the accurate one in case of overlapping of some parts or inconsistent curvature of straight walls;
    \item and the amount of enclosed areas, which represents the same idea: if there are overlapping rooms or some artifacts in an unknown area this metric can be used to detect them.
\end{itemize}

\subsection{Proportion} 
\begin{figure}[h]
	\subfloat[][With blurry effect\label{fig:blurry_effect} \\ ]{\includegraphics[width=0.225\textwidth]{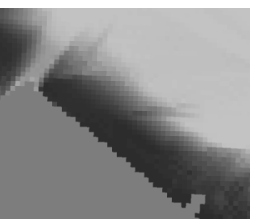} \hspace{2mm}}
	\subfloat[][Without blurry effect \label{fig:blurry_effect_off} \\ ]{\hspace{2mm} \includegraphics[width=0.225\textwidth]{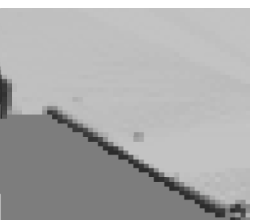}}
	
    \caption{Wall representation}
	\label{fig:wall_representation}
\end{figure}

One of the most perceptible feature of the picture of a map is the accuracy of the walls. With two maps that are similar by most of other common standards, the more blurriness in a map, the lower its quality. In Fig.~\ref{fig:wall_representation} the same wall is presented with high and low blur. 

To determine the amount of blurry effect, one must calculate the proportions of pixels that correspond to walls and to free spaces. Extracting occupied cells might entail difficulties in implementation for maps that contain blurred walls because it is a challenge to distinguish a wall from a free space.

The easiest solution is to consider any cell with probability greater than zero as a wall, but that might lead to the appearance of artifacts in free space. The next idea is to determine a threshold for distinguishing free cells from occupied cells. The threshold should not be hardcoded and it should be customizable for each map, i.e., it should depend on the highest probability of cells, amount of wall blur etc. In this paper the mean value of all cells is used to determine a threshold. All cells whose probabilities of being occupied fall below this threshold should be considered free, and all others are classified as occupied.

The proportion of occupied cells corresponds to the quality of the map: the higher this proportion - the less suitable the algorithm. It was mentioned above that this metric allows a user to find out if wall blur exists. Another use is detection of duplication of a wall on a map. When one sees two very close parallel lines on a map, it could be claimed that most probably this must be the same line, and this metrics allows one to determine whether this type of error has appeared on a map. 

This idea works correctly in assumption that the built map is close to the ground truth, i.e., the estimated map and the ground truth should a similar structure. If this assumption is not fulfilled, using this metric might yield incorrect results. For example, the algorithm lose data and create an incomplete map, in which  case the proportion of occupied cell might be very small. That's why these metrics cannot be used independently of one another and should be used only as a part of a complex analysis.

\subsection{Corner Count}

It is also believed that, with the same sequence, the more corners the estimated map has, the higher the chance that this map is less consistent and has more artifacts. 

For two maps of similar quality generated by two SLAM algorithms on the same sequence, the map with fewer corners is more likely to be more consistent and have fewer artifacts. As long as no information is missed or dropped by any algorithm, and actual corners are successfully reflected in each map, then any extra corners would be errors: artifacts, doubled walls caused by trajectory mismatch, and broken circle areas. Admittedly, different level of details between different SLAM algorithms will make this comparison less convincing, but by applying a stronger rule to define what a "corner" is, this metric could successfully indicate consistency with a reasonable chance empirically. 

For this specific purpose, only structural corners are taken into our account by ignoring small dots sparsely spread throughout the free area. These dots might be small obstacles or just noise, and by ignoring them we  get a more clear and accurate result. 

To extract corners from the raw maps, the map's pixel values are first remapped using the rule that the higher the cell value is, the higher the chance that the cell is not occupied (while unknown area should be zero).  Then a Gaussian-Laplace filter is applied to extract the abstracted structure of the new map. Based on the abstract structure, after dropping small discrete dots (or dot groups), Harris corner detector \cite{harris1988combine} is applied on the preprocessed map structure to get the structural corners in the map.

The number of corners in a map is counted and considered as an indicator of map quality. As mentioned before, this metric is consistent and reasonable if the maps to be compared are at least of similar quality. Counterexamples including an empty map with zero corners could be constructed where this metric could not reflect the map quality at all, but given the assumption of no ground truth supplied, we do expect the maps to have at least similar quality to conduct the comparison, and the shortcoming of this metric is not only unavoidable without ground truth, but also acceptable in practice.

\subsection{Enclosed Areas} 
Another characteristic that can be used to distinguish between low-quality and high-quality maps of the same sequence is the number of enclosed areas in the map. An enclosed area is a region in the map which is bordered completely by cells whose probabilities of being occupied are much higher or unknown. Such bordering cells can be thought of as the walls surrounding an open space.

There are several situations in which the presence of such areas indicates a failure. For example, when a room is scanned multiple times but not recognized, so that the final map is composed of slight rotations of the room superimposed upon each other, there are multiple enclosed triangles on the outer edges of the explored region. Another possibility is failure of loop closure, when a robot fails to recognize that it has returned to the same place where it started. In that case, one might see an overlap between the first and last parts of the map constructed.

\begin{figure}[h]
    \centering
    \includegraphics[width=0.3\textwidth]{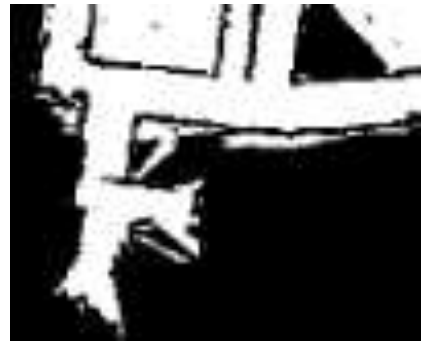}
    \caption{The part of a map with several enclosed areas}
    \label{fig:enclosed_areas_example}
\end{figure}

To obtain a useful map, one must divide all cells into two categories: occupied/undefined, and free. The pixel values of undefined regions are remapped to the value corresponding to the maximum probability of being occupied. Next, the map is reduced to a binary image using Otsu's method \cite{otsu1979threshold}. After thresholding, occupied and undefined regions should have the same value. Enclosed areas are found with Suzuki's algorithm \cite{suzuki1985topological}, and the process is repeated, with undefined areas being reassigned a value which corresponds to a slightly lower probability of being occupied. Thus one iterates through the possible values of undefined regions, and returns the maximum number of enclosed areas found. The example of extracted enclosed areas is shown in Fig~\ref{fig:enclosed_areas_example}.

In cases where there is not a wide spread between the values of the cells most likely to be occupied and those of the cells least likely to be occupied (i.e., the borders are not sharp), this approach may lead to an overestimate of the number of enclosed areas. However, the ambiguity itself is an indication of a problem, and the more of these there are, the lower the quality of the map. This metric works best when similar kinds of failures occur in map construction, as it assumes that valid enclosed areas are detected in both maps, and the difference between counts should only consist of problematic areas. Obviously, it would fail to pick up that there is anything wrong with a completely blank map. But as with the metric described above, the assumptions made to use this metric are acceptable in practice, and necessary when no ground truth is available.

\section{Testing service}

\subsection{Solution description}

The evaluation methods that are mentioned in Sec~\ref{sec:Metrics} are implemented in service created by the authors. This service allows the user to execute several SLAM algorithms on downloaded data sequences and provides numerical and qualitative results of evaluation.

The set of SLAM algorithms is chosen by authors and will be expanded in the future. The current set includes gmapping~\cite{gmapping}, an implementation of tinySLAM~\cite{huletski2016tinyslam}, vinySLAM~\cite{vinySLAM}, and Google’s algorithm cartographer~\cite{hess2016real}. This list involves the referee algorithm - gmapping, one novel algorithm - cartographer, and algorithms from SLAM constructor framework for ROS~\cite{huletski2016slam}. The algorithms from framework are chosen because they are created in the authors laboratory, so evaluation results got with testing service could define quality of SLAM framework algorithms in comparison with judge ones.

To evaluate algorithms, their ROS~\cite{o2014gentle} implementations are used. This operation system was chosen because it presents a unified way to launch various algorithms. Moreover, it is a popular base for modern SLAM algorithms and many of them comes are included in the ROS API. ROS also makes it possible to provide data and collect results in a structured format.

MIT sequences~\cite{fallon2013stata} are currently used as input data. This dataset is supported with a ground truth trajectory, so one can apply trajectory RMSE value to estimate an algorithm result quantitatively. In the future, more datasets will be added, like those of Deutsches Museum~\cite{deutsches-museum} or Willow Garage~\cite{mason2012object}. Unfortunately, the mentioned datasets are not provided with ground truth trajectories or ground truth maps, so the amount of methods that could be used for evaluation of SLAM algorithms is reduced. Datasets are often presented in different formats, so it is another task to convert them to a unified structure. 

Generally, the testing service requires the user to choose a SLAM algorithm (or several algorithms), a sequence (or sequences), and the number of iterations for which an algorithms will run. There is one another field that could be filled - extra parameters field. The testing service allows to configure variables for SLAM algorithms. There could be variables for a scan matching process, for a map view etc. Thus, one can evaluate SLAM approaches under similar conditions.

To present a user interface for testing service, a Jenkins framework was chosen.

Every evaluation launch is carried out on a remote server so that it does not take up user computation resources. The iterations are split  into parallel launches, so that the evaluation time is reduced. Every launch requires all server resources, and the next execution can be carried out only after all previous builds have finished.

For every SLAM algorithm and every sequence the following output files are created:
\begin{itemize}
    \item a text file with the RMSE mean and standard deviation in meters;
    \item PGM representations of a map at every stage as it is built by the SLAM algorithm (including a GIF file as an union of all these pictures);
\item a text file with the amount of corners extracted from a map;
    \item a text file with the ratio of occupied cells and to all cells;
    \item a text file with the pose log of the mobile platform.
\end{itemize}

A utility which extracts information from all these files and displays it as an html page. It presents a table with RMSE values only if a data sequence is provided with a ground truth trajectory. This table is easy to analyze because a lower value of RMSE means that the corresponding SLAM algorithm provide better results. The RMSE value is consistent by itself, not only relative to the RMSE values of other SLAM algorithms. An example of a table filled with RMSE values of trajectories that are obtained from the previously mentioned SLAM algorithms is presented in Tab~\ref{tab:RMSE}.

\begin{figure*}[h]
	\centering
	\null\hfill
	\subfloat[][Cartographer\label{fig:Cart_01-28} \\ ]{\includegraphics[height=0.2\textheight]{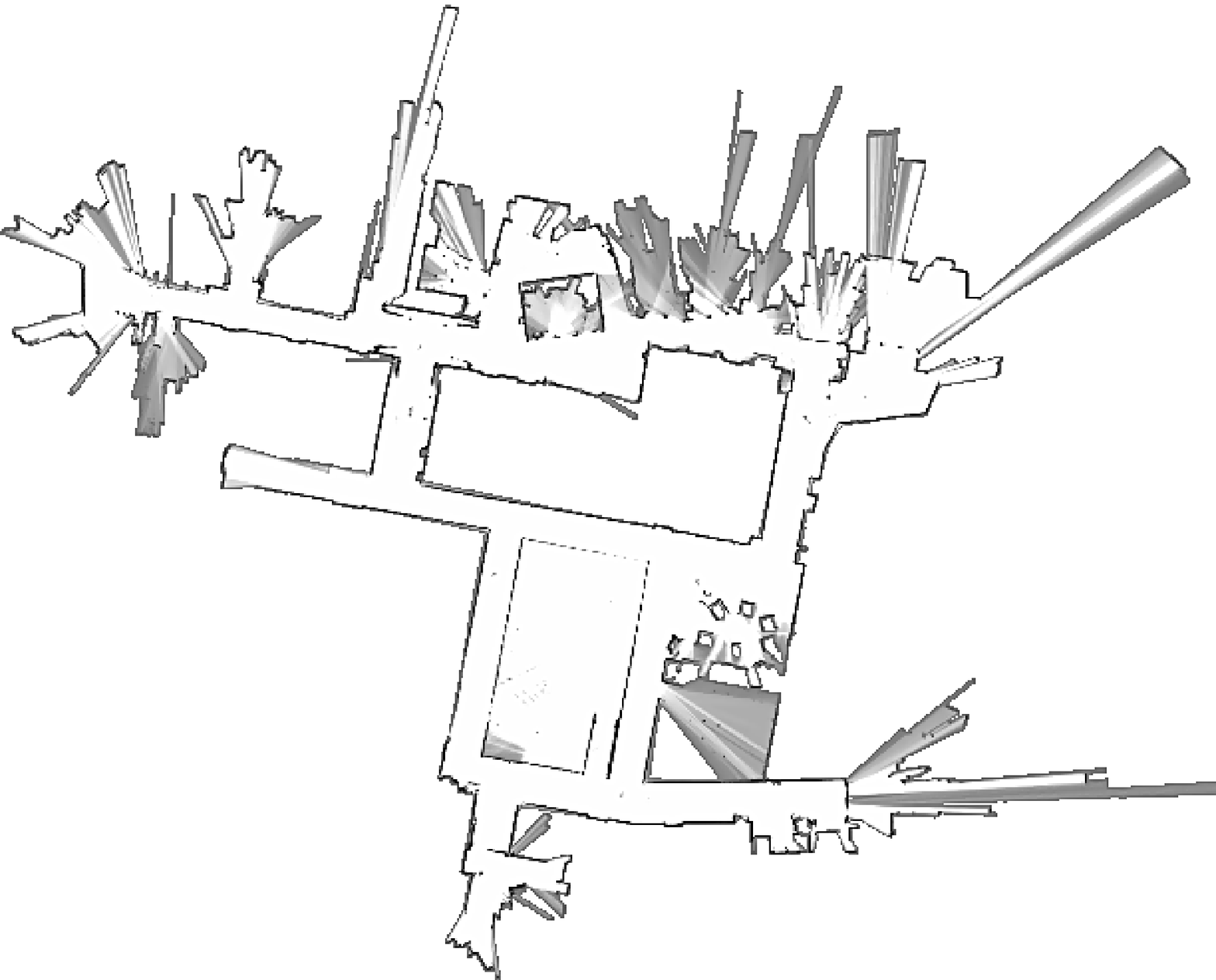}}
	\hfill
	\subfloat[][gmapping\label{fig:gmg_01-28} \\ ]{\includegraphics[height=0.2\textheight]{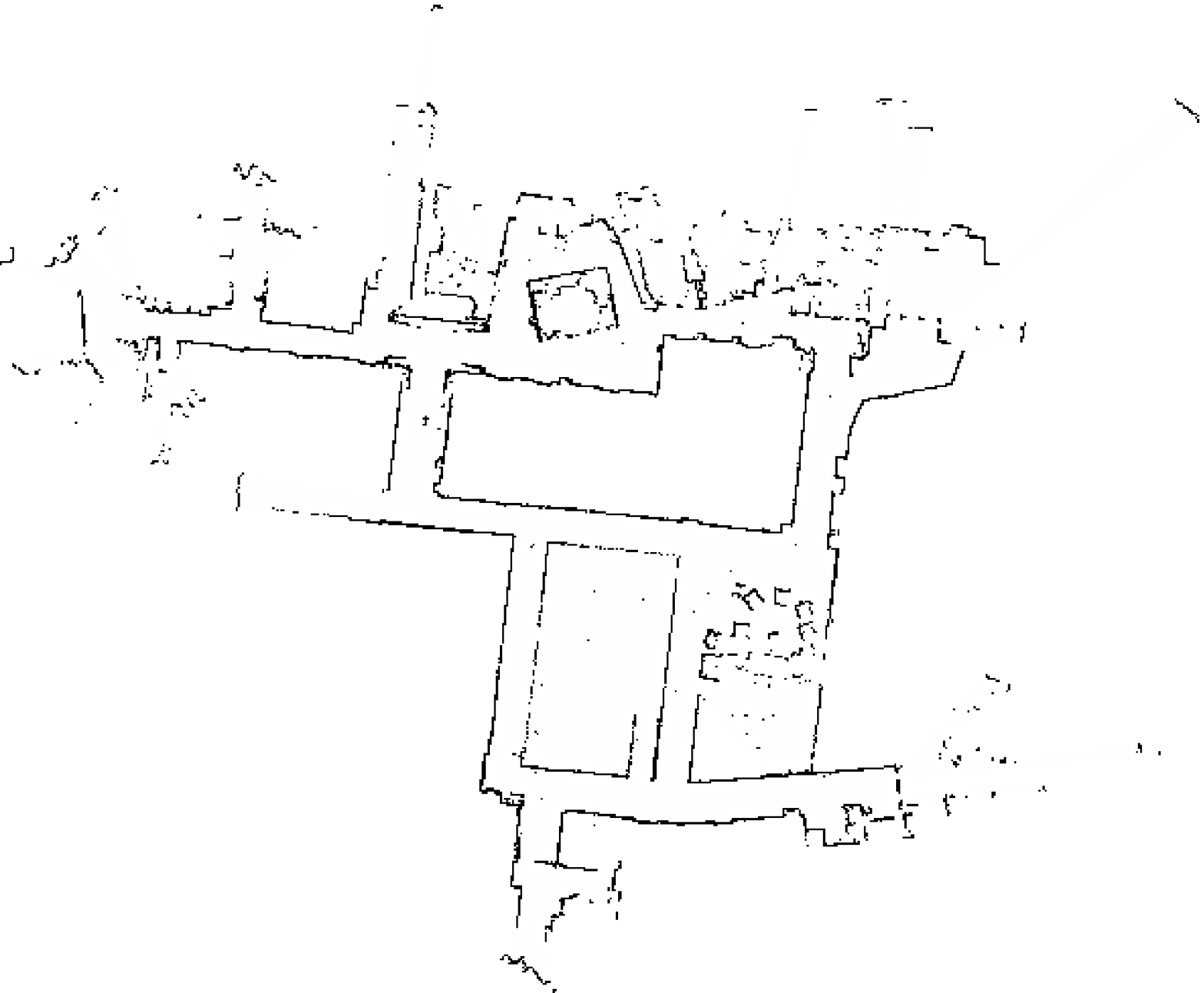}}
	\hfill
	\subfloat[][hectorSLAM\label{fig:hec_01-28} \\ ]{\includegraphics[height=0.2\textheight]{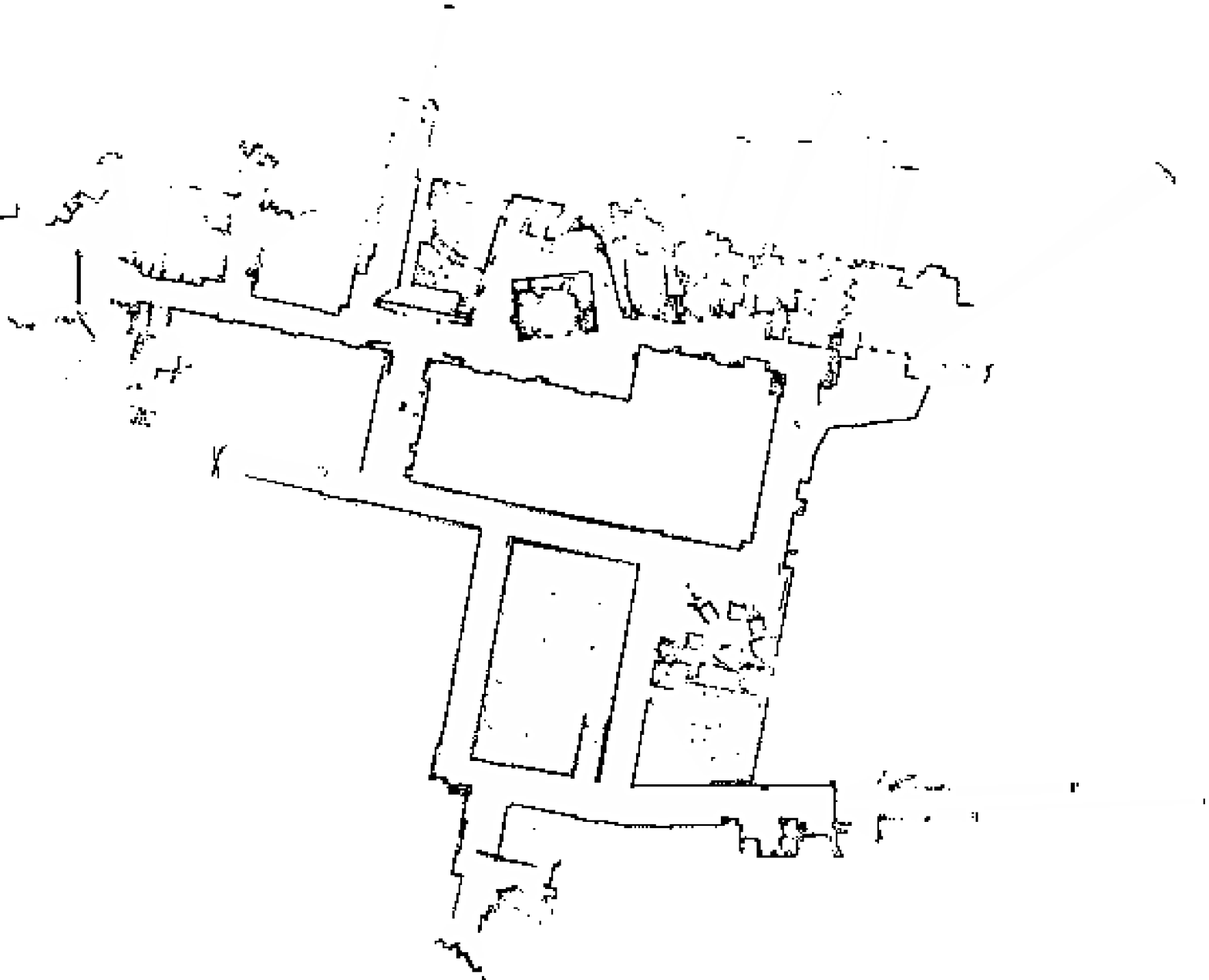}}
	\hfill\null
	\hfil

	\null\hfill
	\subfloat[][tinySLAM(worst)\label{fig:tiny_01-28_w} \\ ]{\includegraphics[height=0.15\textheight]{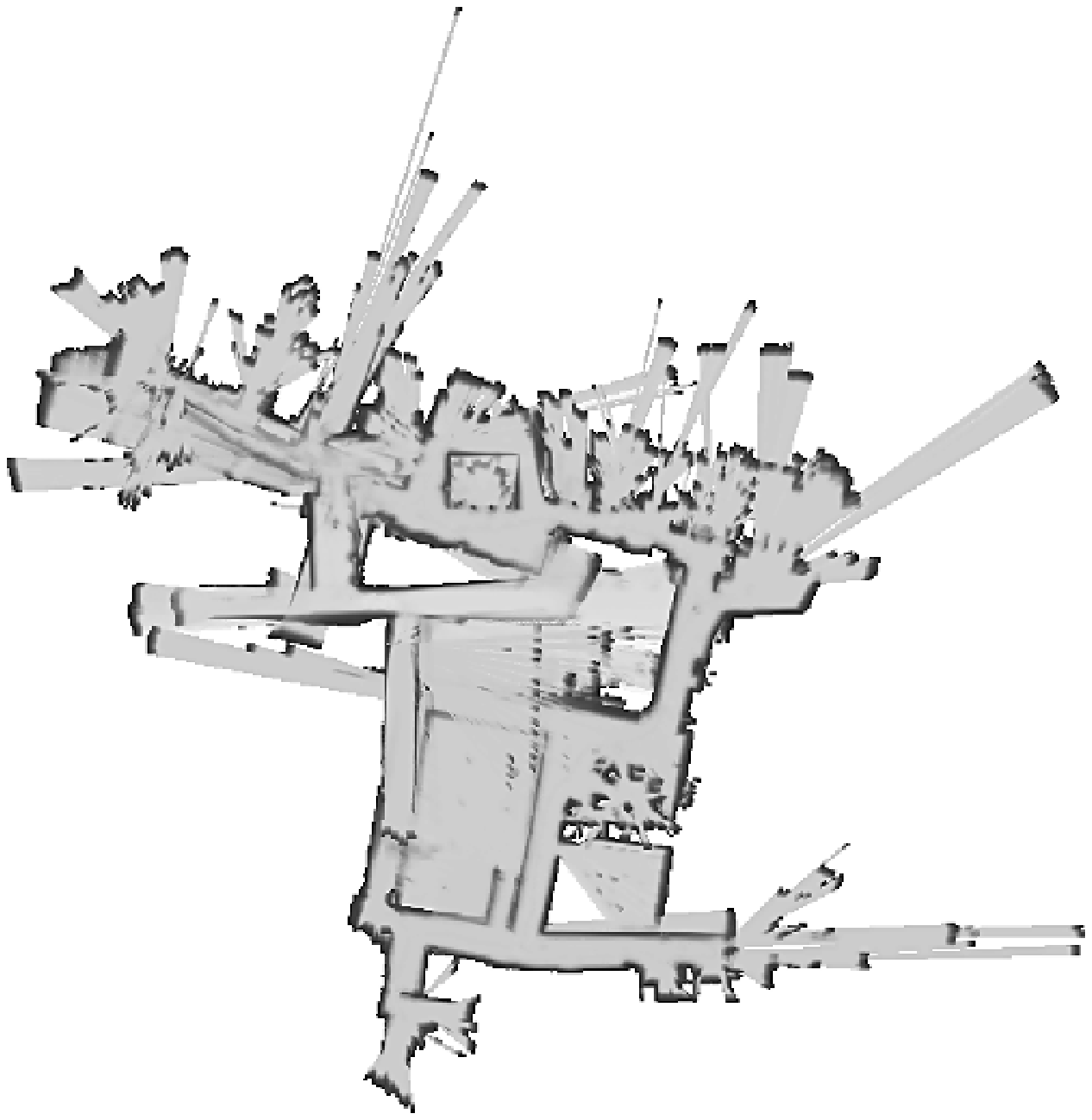}}
	\hfill
	\subfloat[][tinySLAM(best)\label{fig:tiny_01-28_b} \\ ]{\includegraphics[height=0.15\textheight]{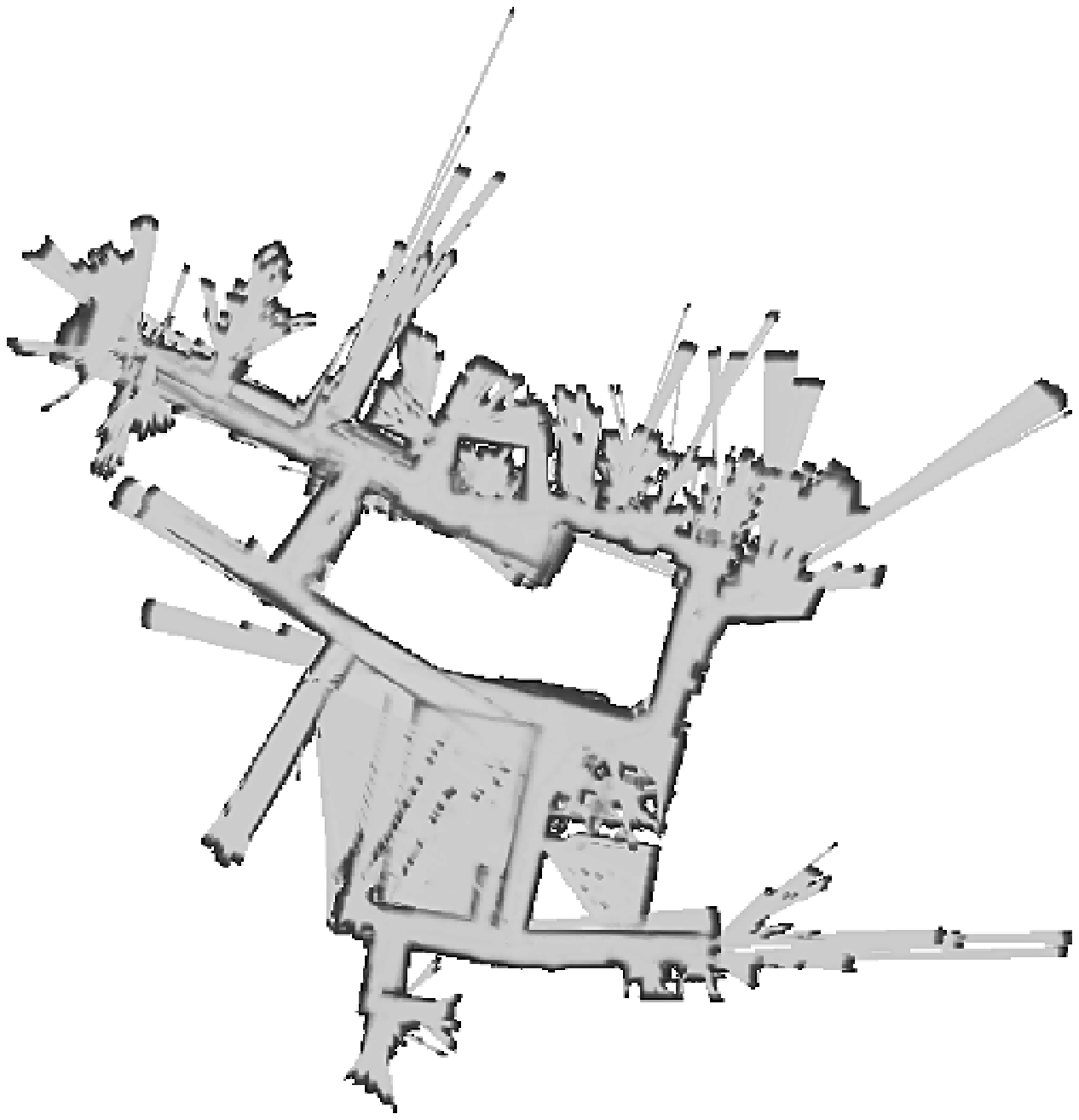}}
	\hfill
	\subfloat[][vinySLAM(worst)\label{fig:viny_01-28_w} \\ ]{\includegraphics[height=0.15\textheight]{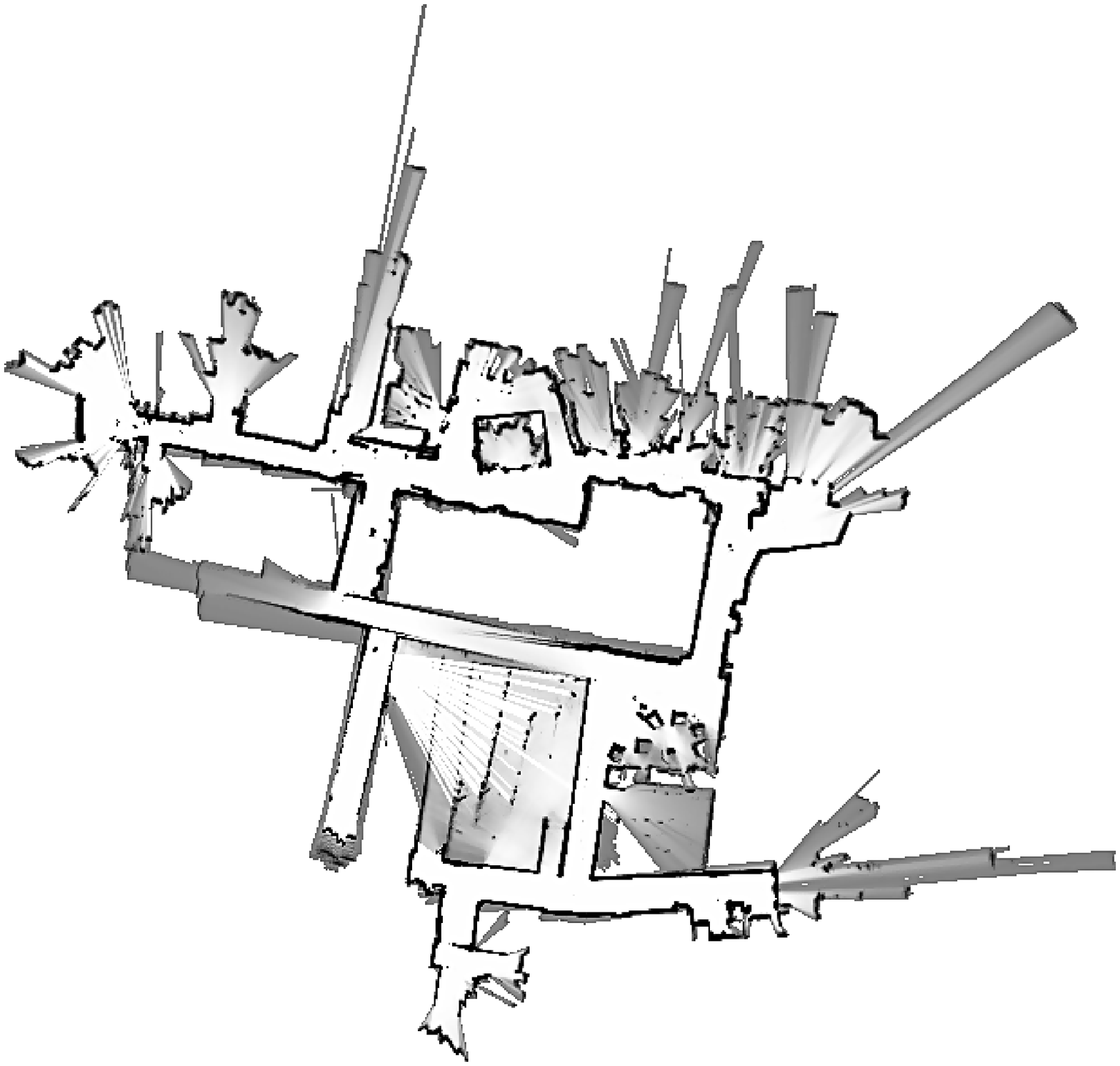}}
	\hfill
	\subfloat[][vinySLAM(best)\label{fig:viny_01-28_b} \\ ]{\includegraphics[height=0.15\textheight]{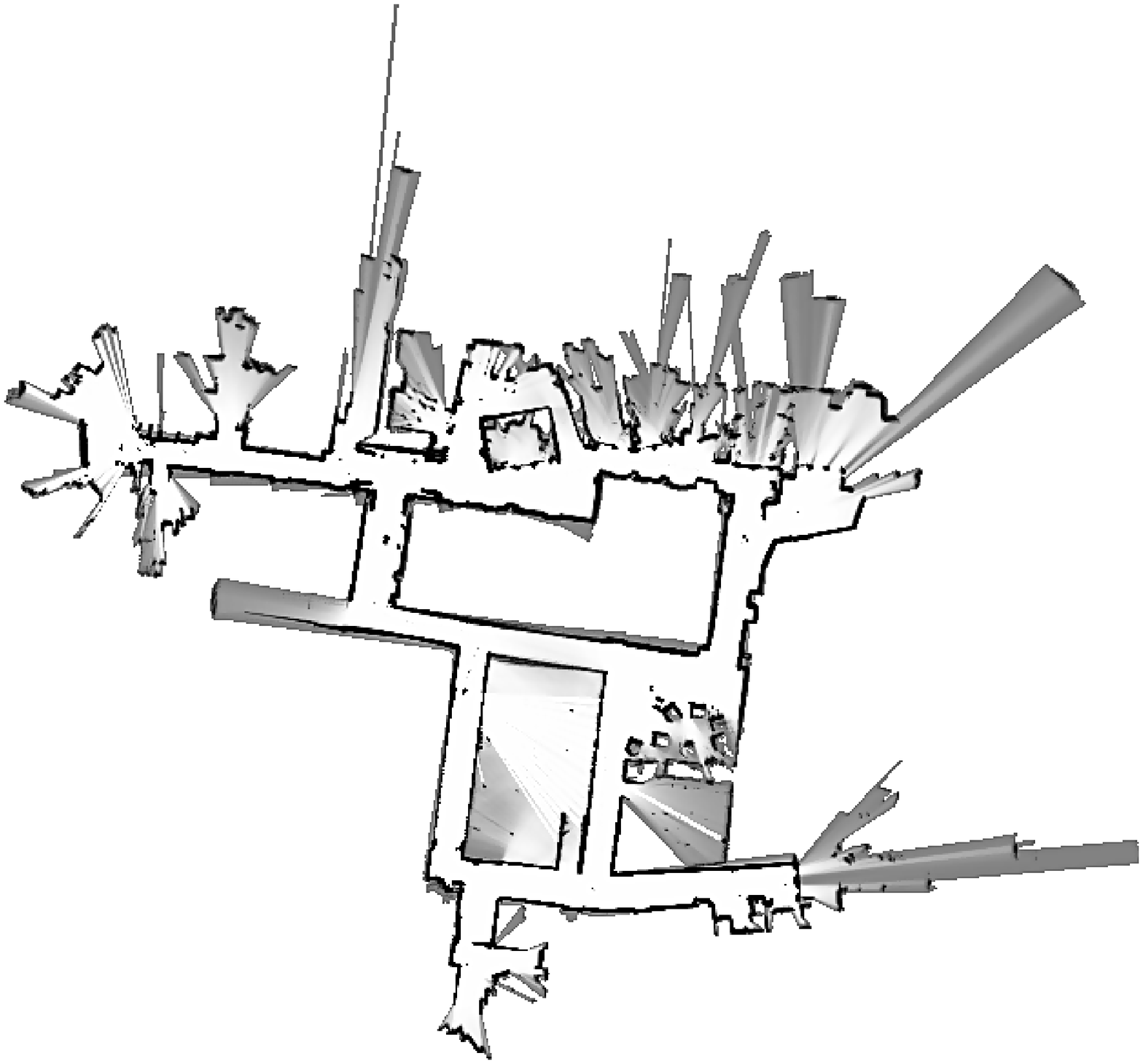}}
	\hfill\null
	\hfil
	
	\caption{Maps of 2011-01-28-06-37-23 MIT sequence built with different SLAMs}
	\label{fig:maps_01-28}
\end{figure*}

Another table that generated by the utility consists of a set of maps - pictures that illustrate a result views of an environment got from launched SLAMs. This evaluation is not quantitative, but it allows the user to recognize cases when one SLAM algorithm provides better estimations than another. For example, an algorithm could work perfectly in a corridor environment but fail when provided data which includes huge measurement errors. An example of a table is presented on a Fig~\ref{fig:maps_01-28}, Fig.~\ref{fig:maps_03-11}.

Moreover, there are three more figures with quantitative results, that presents amount of extracted corners, amount of enclosed areas, and a ratio between amount of undefined cell and total amount of cells on maps. These metrics are relative and could be used for evaluation only when several SLAM algorithms are launched on the same data sequence. On the other hand these, metrics do not require a ground truth map or trajectory. The mentioned values are presented in Fig~\ref{fig:Wall_proportion}, Fig~\ref{fig:Corners} and Fig~\ref{fig:Enclosed_areas}.

\subsection{Experimental results}

\begin{figure}[h]
    \centering
    \includegraphics[width=0.5\textwidth]{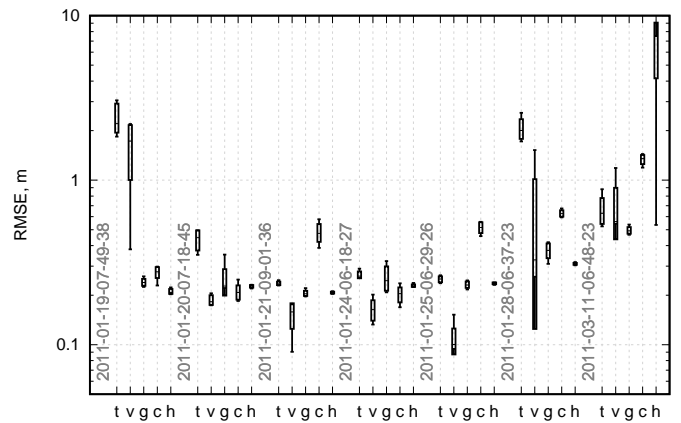}
    \caption{Trajectory RMSE values}
    \label{fig:RMSE}
\end{figure}

\begin{figure*}[h]
	\centering
	\null\hfill
	\subfloat[][Cartographer(worst)\label{fig:Cart_03-11_w} \\ ]{\includegraphics[height=0.12\textheight]{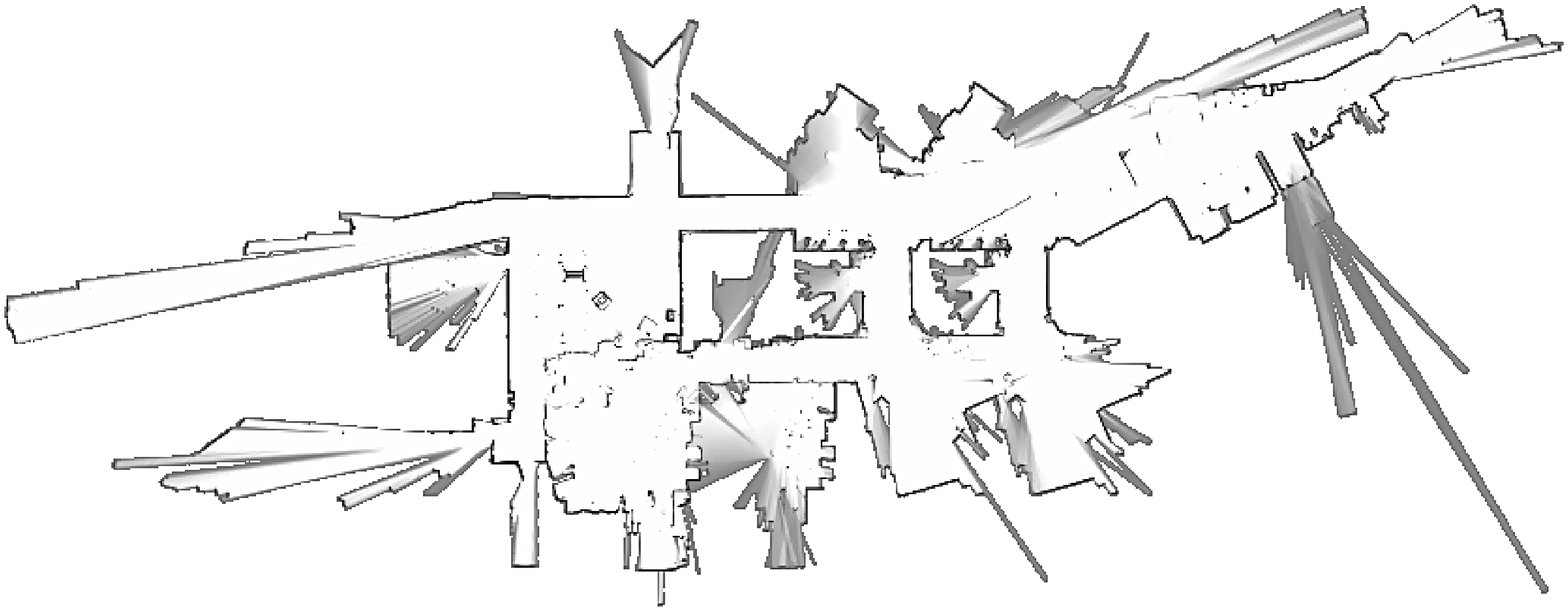}}
	\hfill
    \subfloat[][Cartographer(best)\label{fig:Cart_03-11_b} \\ ]{\includegraphics[height=0.12\textheight]{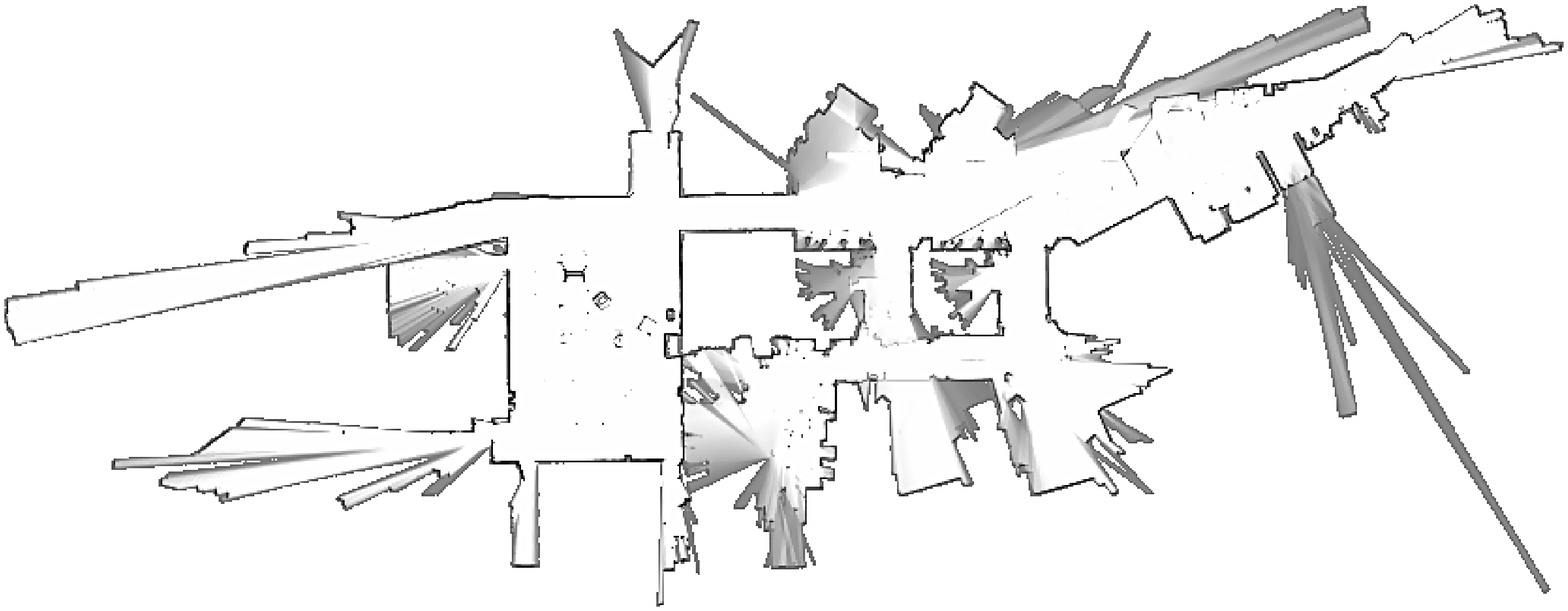}}
	\hfill\null
	\hfill
	
	\null\hfill
	\subfloat[][hectorSLAM(worst)\label{fig:hec_03-11_w} \\ ]{\includegraphics[height=0.12\textheight]{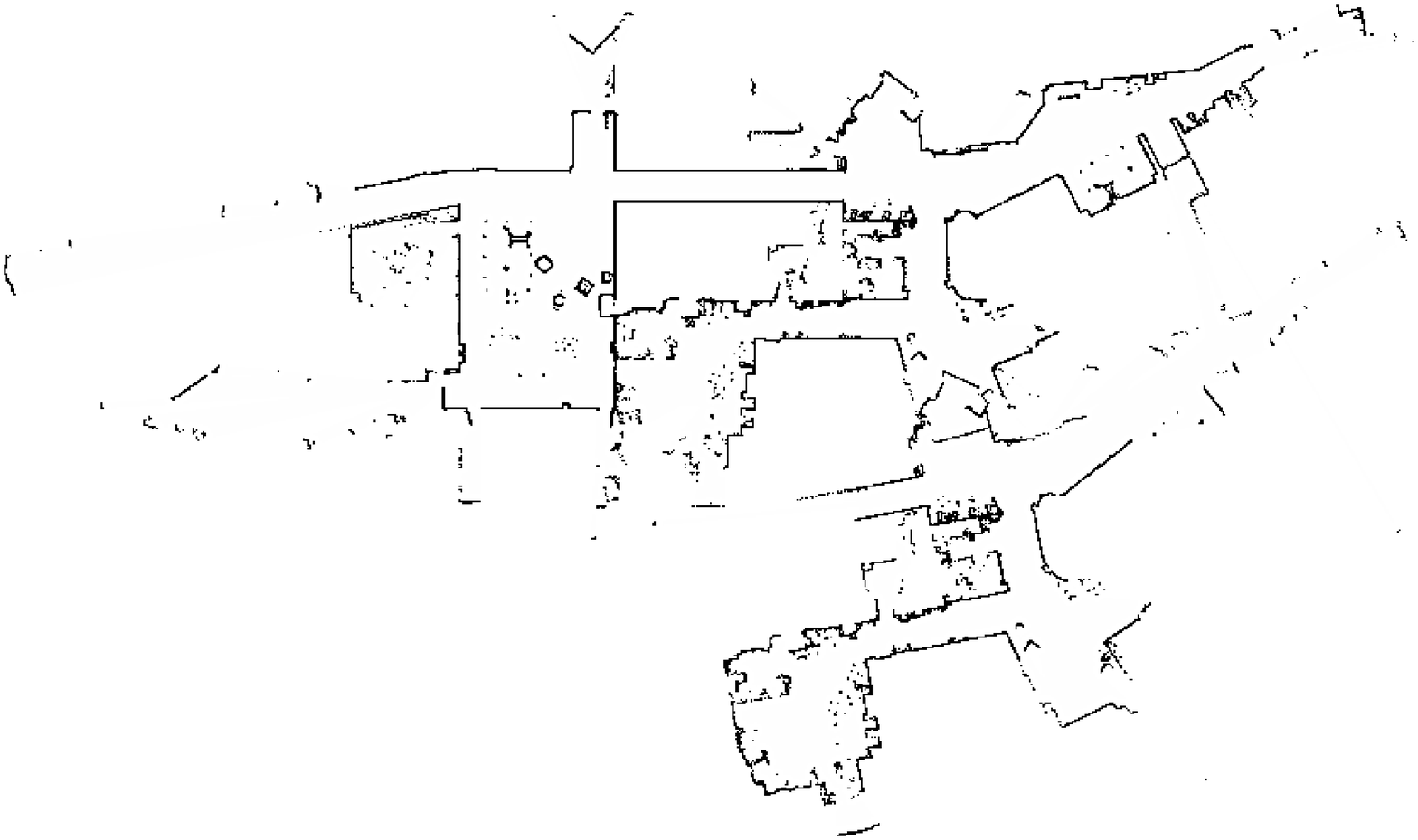}}
	\hfill
	\subfloat[][hectorSLAM(best)\label{fig:hec_03-11_b} \\ ]{\includegraphics[height=0.12\textheight]{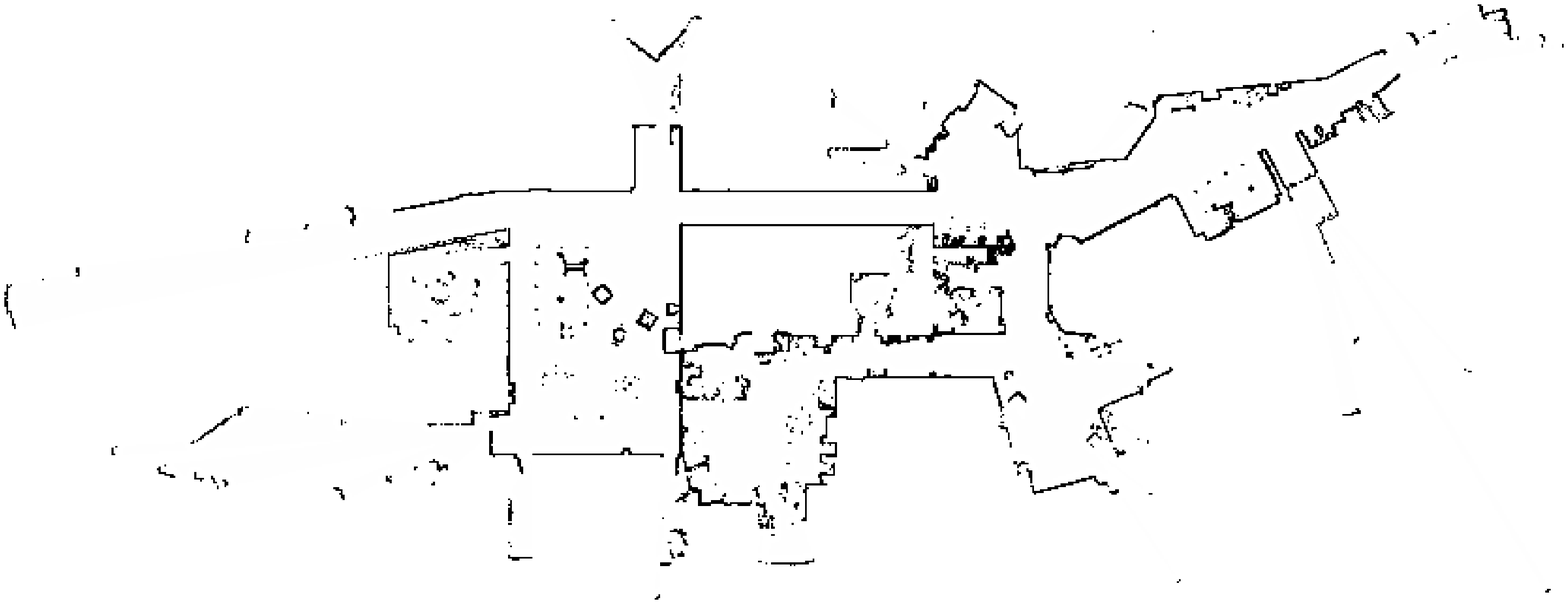}}
	\hfill\null
	\hfil

	\null\hfill
	\subfloat[][tinySLAM(worst)\label{fig:tiny_03-11_w} \\ ]{\includegraphics[height=0.12\textheight]{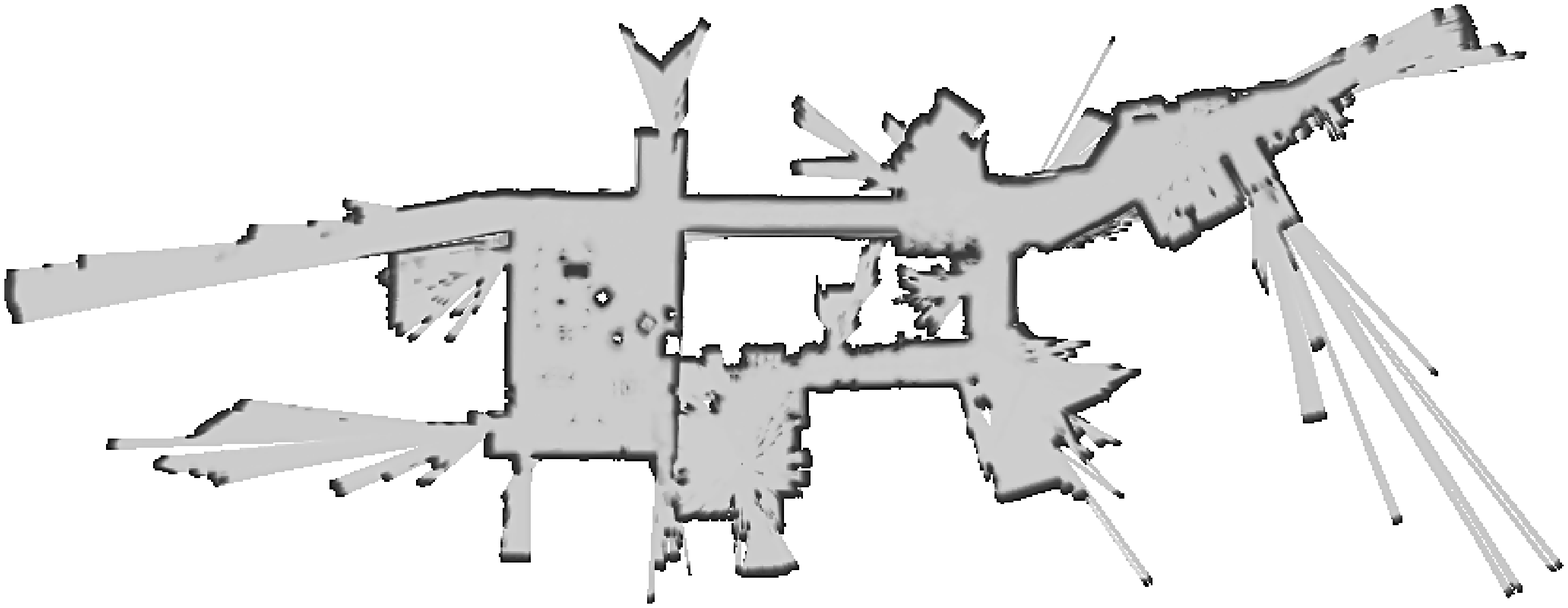}}
	\hfill
	\subfloat[][tinySLAM(best)\label{fig:tiny_03-11_b} \\ ]{\includegraphics[height=0.12\textheight]{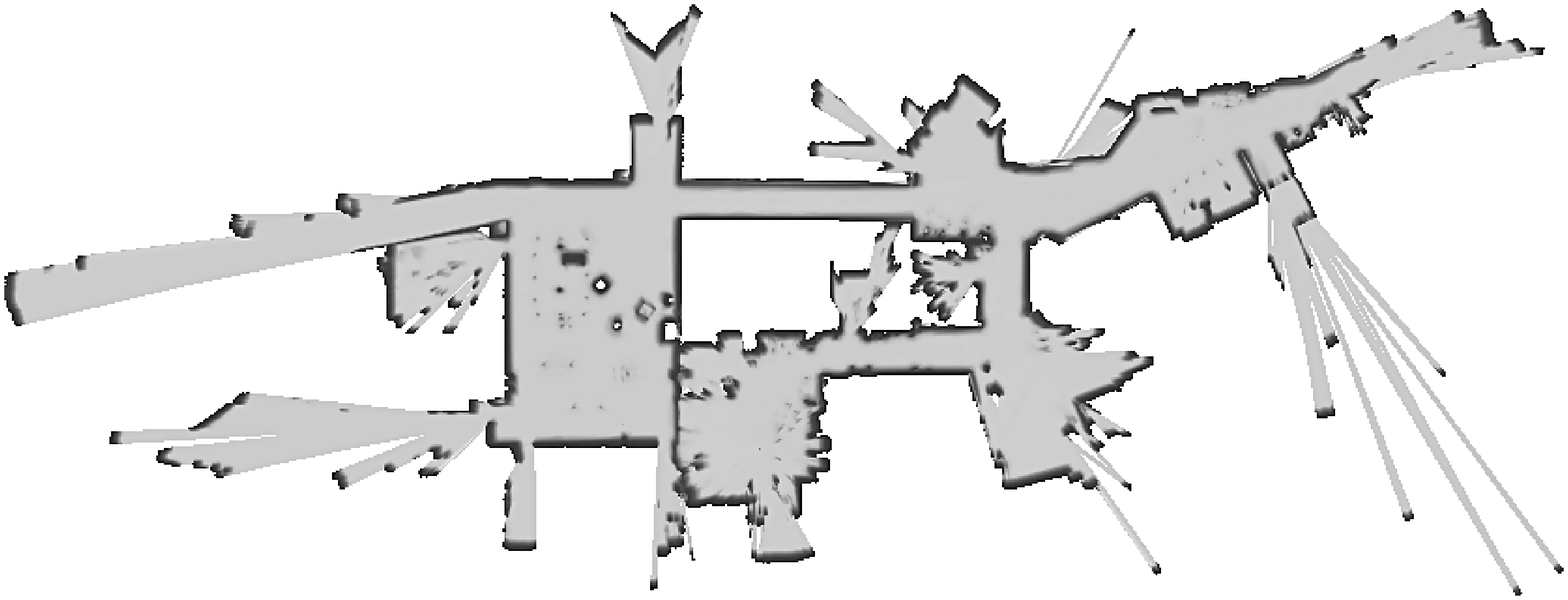}}
	\hfill\null
	\hfil
	
	\null\hfill
	\subfloat[][vinySLAM(worst)\label{fig:tiny_03-11_w} \\ ]{\includegraphics[height=0.12\textheight]{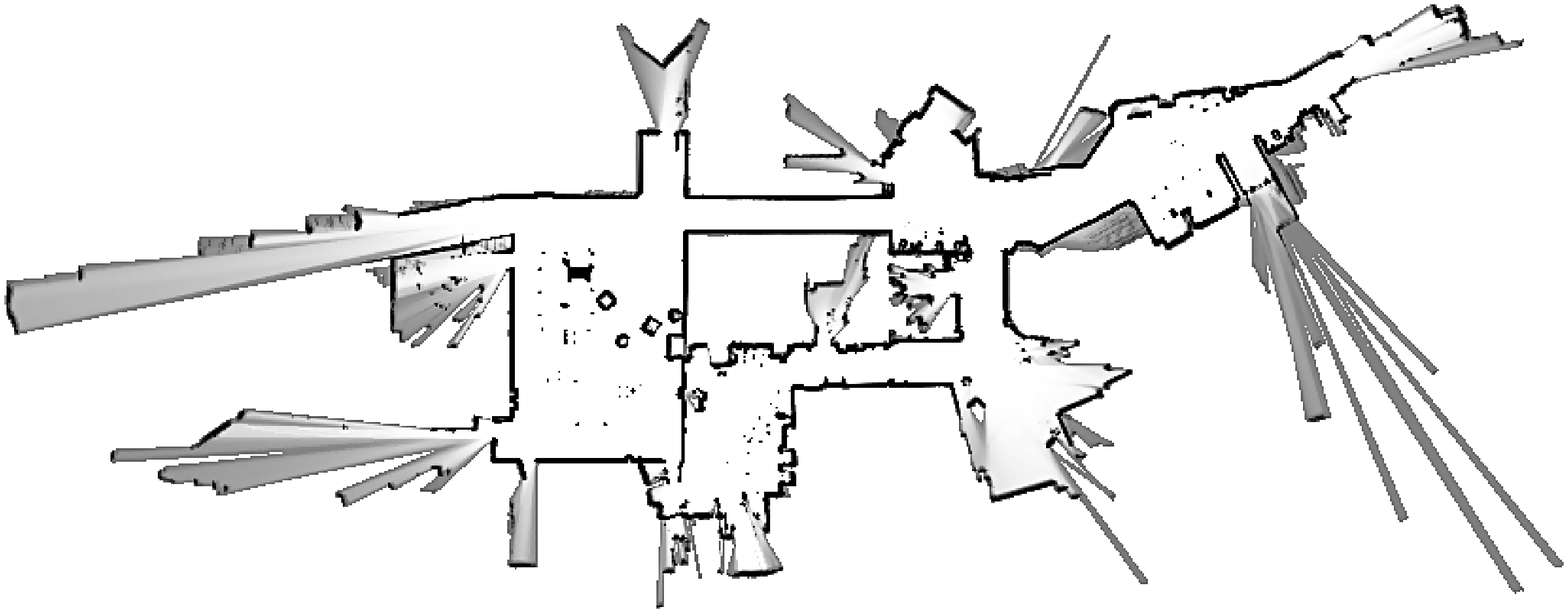}}
	\hfill
	\subfloat[][vinySLAM(best)\label{fig:tiny_03-11_b} \\ ]{\includegraphics[height=0.12\textheight]{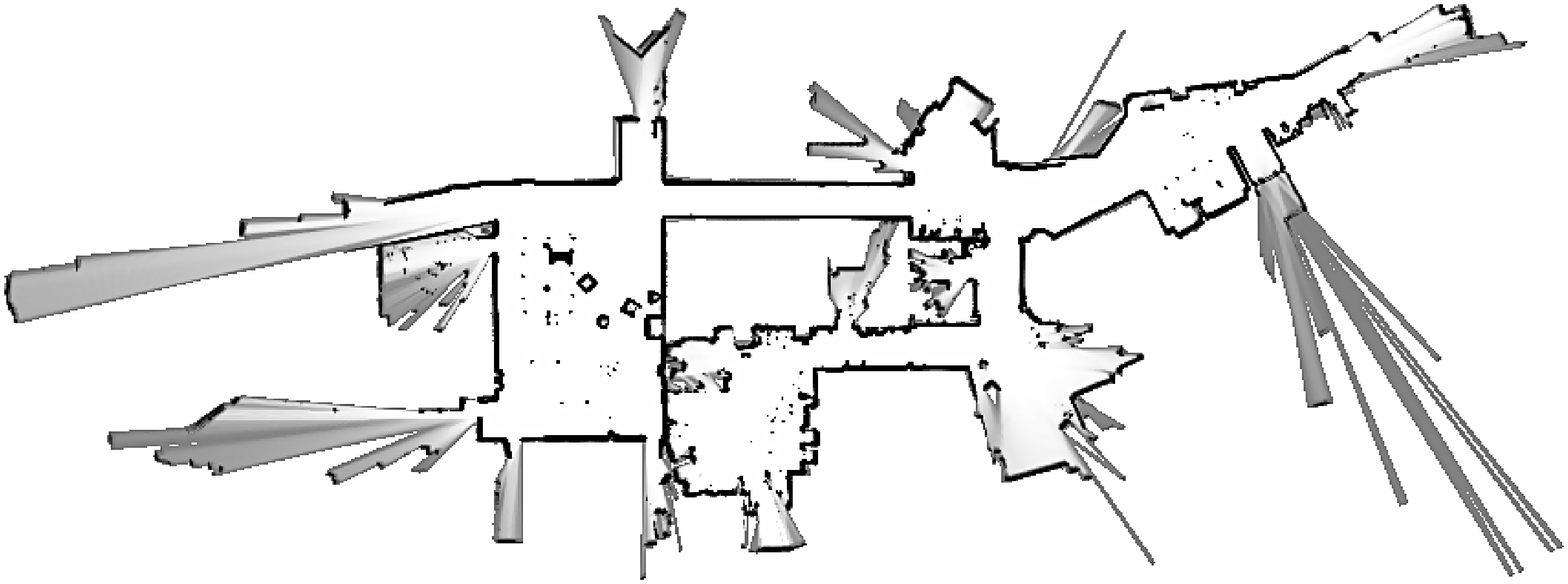}}
	\hfill\null
	\hfil
	
	\caption{Maps of 2011-03-11-06-48-23 MIT sequence built with different SLAMs}
	\label{fig:maps_03-11}
\end{figure*}

The experimental results are presented in tables and figures below. There evaluation was conducted on MIT data sequences~\cite{fallon2013stata} that have different length and structure of environment (different amount of rooms and corridors). The trajectory RMSE values are presented in a Tab~\ref{tab:RMSE} and in a Fig~\ref{fig:RMSE}. The figure has the logarithmic scale of ordinate for better matching a relation between SLAM algorithms output results. These results show that gmapping, Cartographer and hectorSLAM are the most robust (have the lowest dispersion) but, for example, hectorSLAM or Cartographer may fail on some sequences and create inconsistent map. The example of such execution is presented on a Fig~\ref{fig:hec_03-11_w}. To recognize conditions when this may happen the GIF file is provided by the testing service.

\begin{table*}[h]
\centering
\caption{RMSE values (m) for evaluated SLAM algorithms}
\label{tab:RMSE}
\begin{tabular}{|c|c|c|c|c|c|c|}
\hline
\multirow{2}{*}{The sequence} & \multirow{2}{*}{Dist, m} & \multicolumn{5}{c|}{SLAM algorithms}                                                 \\
\cline{3-7} \rule{0cm}{0.28cm}
                              &                          & tinySLAM          & vinySLAM          & gmapping          & cartographer      & hectorSLAM \\
\hline
\rule{0cm}{0.28cm}
2011-01-19-07-49-38           & 68                       & 2.432 $\pm$ 0.488 & 1.581 $\pm$ 0.579 & 0.239 $\pm$ 0.011 & 0.275 $\pm$ 0.021 & 0.211 $\pm$ 0.007\\
\hline
\rule{0cm}{0.28cm}
2011-01-20-07-18-45           & 76                       & 0.434 $\pm$ 0.061 & 0.187 $\pm$ 0.012 & 0.243 $\pm$ 0.044 & 0.208 $\pm$ 0.021 & 0.226 $\pm$ 0.004\\
\hline
\rule{0cm}{0.28cm}
2011-01-21-09-01-36           & 87                       & 0.235 $\pm$ 0.006 & 0.151 $\pm$ 0.027 & 0.205 $\pm$ 0.007 & 0.482 $\pm$ 0.059 & 0.207 $\pm$ 0.003\\
\hline
\rule{0cm}{0.28cm}
2011-01-24-06-18-27           & 87                       & 0.266 $\pm$ 0.013 & 0.162 $\pm$ 0.023 & 0.256 $\pm$ 0.043 & 0.202 $\pm$ 0.022 & 0.228 $\pm$ 0.004\\
\hline
\rule{0cm}{0.28cm}
2011-01-25-06-29-26           & 109                      & 0.249 $\pm$ 0.011 & 0.106 $\pm$ 0.019 & 0.231 $\pm$ 0.011 & 0.515 $\pm$ 0.039 & 0.236 $\pm$ 0.003\\
\hline
\rule{0cm}{0.28cm}
2011-01-28-06-37-23           & 145                      & 2.064 $\pm$ 0.282 & 0.569 $\pm$ 0.444 & 0.374 $\pm$ 0.038 & 0.628 $\pm$ 0.028 & 0.311 $\pm$ 0.004\\
\hline
\rule{0cm}{0.28cm}
2011-03-11-06-48-23           & 245                      & 0.660 $\pm$ 0.118 & 0.667 $\pm$ 0.231 & 0.496 $\pm$ 0.022 & 1.336 $\pm$ 0.086 & 6.611 $\pm$ 2.455\\
\hline
\end{tabular}
\end{table*}

\begin{figure}[h]
    \centering
    \includegraphics[width=0.5\textwidth]{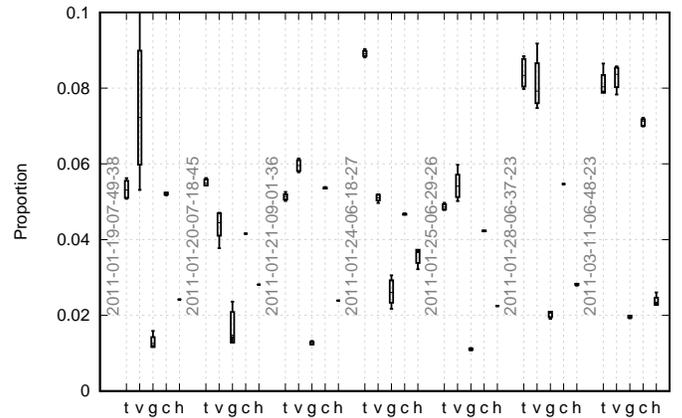}
    \caption{The proportion of occupied cells to the sum of free and unknown}
    \label{fig:Wall_proportion}
\end{figure}

The proportion of occupied cells that was described in Sec~\ref{sec:Metrics} are shown in a Fig~\ref{fig:Wall_proportion}. It presents that tinySLAM, vinySLAM and Cartographer have blurry effect and gmapping and hectorSLAM don't. Following the description of this metric, the most accurate algorithm is gmapping but it is early to draw conclusions about the precision of all algorithms. However one can see that vinySLAM usually has the valuable dispersion of the considered proportion.

\begin{figure}[h]
    \centering
    \includegraphics[width=0.5\textwidth]{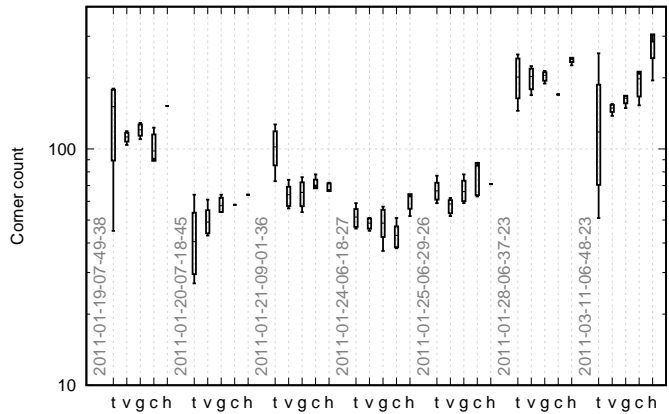}
    \caption{The amount of corners extracted from maps}
    \label{fig:Corners}
\end{figure}

Amount of corners for each algorithm are presented in a Fig~\ref{fig:Corners}. This figure has logarithmic scale of vertical axis of corner amount. This metric shows that vinySLAM usually works better than Cartographer and hectorSLAM, and moreover has a little dispersion. This means that maps built by vinySLAM have more accurate walls and less duplication of corners. Moreover vinySLAM shows close and sometimes better results to gmapping and this means that it is one of the most accurate algorithms in terms of this metric. At the same time tinySLAM has the greatest dispersion because it uses the math apparatus based on the random and, therefore, this algorithm cannot be estimated with this metric accurately. 

\begin{figure}[h]
    \centering
    \includegraphics[width=0.5\textwidth]{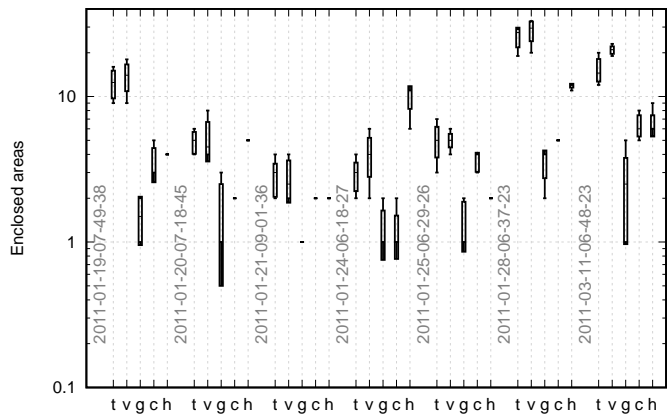}
    \caption{The amount of enclosed areas for SLAM algorithms}
    \label{fig:Enclosed_areas}
\end{figure}

The amount of enclosed areas is the most sensitive metric to the random parts of algorithm as it is presented in Fig~\ref{fig:Enclosed_areas}. In this figure the ordinate axis has also logarithmic scale. The huge advantage of gmapping is explained by the little proportion of occupied cells in this algorithm: several areas that could be counted as enclosed were not considered because of the absence of border. On the other hand the previous metrics show that gmapping is very accurate algorithm and the results in this figure prove this fact.

Looking at the performed results it is possible to conclude that gmapping, Cartographer and vinySLAM are the most accurate of five considered algorithms. Gmapping has got the highest score but it is challenging to determine which algorithm should be in the second place. Definitely the fourth place is occupied by hectorSLAM while tinySLAM is in the last place shows in case of the valuable dispersion of the results.

\section{Conclusion}
In this paper, a framework to quantitatively evaluate SLAM algorithms is presented. The metrics introduced capture characteristics of the maps generated by SLAM algorithms, and on the basis of those characteristics distinguish a high-quality map from a low-quality map of the same sequence. These metrics are well suited to realistic usages of SLAM algorithms, where ground truth maps and trajectories are not always available. Furthermore, they yield other information about map creation that is also important in practical contexts, but for which no quantitative solutions appear to exist: the signaling of possible artifacts and failure in map creation.

\section*{Acknowledgment}
Authors would like to thank JetBrains Research\footnote{JetBrains Research website: \url{https://research.jetbrains.org}} for provided support and materials for working on this paper.

\def\BibTeX{BibTeX}
\bibliographystyle{IEEEtran}
\bibliography{./paper}

\end{document}